%% file: main.tex
\documentclass[12pt,logo]{artechreport}
\usepackage{amsmath,amssymb,amsfonts} 

\usepackage{amssymb}
\usepackage{pifont}

\usepackage{fourier} 
\usepackage{microtype} 
\usepackage{csquotes}
\usepackage{hyperref}      
\usepackage{url}                      
\usepackage{booktabs}                 
\usepackage{amsmath,amssymb,amsfonts} 
\usepackage{nicefrac}                 
\usepackage{microtype}                
\usepackage[numbers,compress]{natbib}
\usepackage{subcaption}
\usepackage{enumitem}
\usepackage{fancyhdr}
\usepackage{graphicx}
\usepackage{wrapfig,booktabs}
\usepackage{xcolor} 
\usepackage{fancyhdr}
\usepackage{blindtext}
\usepackage{marvosym}
\usepackage{tikz}
\usepackage{microtype}
\usepackage{setspace}
\usepackage{wrapfig,lipsum,booktabs}
\usepackage{float}
\usepackage{subcaption}
\usepackage{mdframed}

\usepackage{array}
\newcolumntype{x}[1]{>{\centering\let\newline\\\arraybackslash}m{#1}}
\usepackage{multirow}

\renewcommand{\today}{2025 Aug 11}

\providecommand{\DeepFleet}{\textls[0]{\normalfont {\scshape DeepFleet}}\xspace}

\newcommand{\cmark}{\ding{51}}%
\newcommand{\xmark}{\ding{55}}%

\title{
\DeepFleet: Multi-Agent Foundation Models for Mobile Robots
}

\author{
Amazon Robotics \hyperref[sec:authors]{\includegraphics[height=12pt]{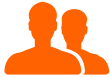}}
}

\begin{document}

\begin{abstract}
\input{abstract.tex}
\end{abstract}

\maketitle

\input{sections/01_Intro}
\input{sections/02_Related_Work}
\input{sections/03_Problem_Formulation}
\input{sections/04_Findings}
\input{sections/05_Robot_Centric_Model}

\input{sections/06_Robot_Floor_Model}
\input{sections/07_Image_Floor_Model}

\input{sections/08_Graph_Floor_Model}
\input{sections/09_conclusion}

\bibliographystyle{unsrtnat}
\bibliography{references}

\newpage 

\input{sections/A_Contributors}

\end{document}

%% file: abstract.tex
We introduce \DeepFleet, a suite of foundation models designed to support coordination and planning for large-scale mobile robot fleets. These models are trained on fleet movement data, including robot positions, goals, and interactions, from hundreds of thousands of robots in Amazon warehouses worldwide. 
\DeepFleet consists of four architectures that each embody a distinct inductive bias and collectively explore key points in the design space for multi-agent foundation models: 
the robot-centric (RC) model is an autoregressive decision transformer operating on neighborhoods of individual robots; the robot-floor (RF) model uses a transformer with cross-attention between robots and the warehouse floor; the image-floor (IF) model applies convolutional encoding to a multi-channel image representation of the full fleet; and the graph-floor (GF) model combines temporal attention with graph neural networks for spatial relationships. 
In this paper, we describe these models and present our evaluation of the impact of these design choices on prediction task performance. We find that the robot-centric and graph-floor models, which both use asynchronous robot state updates and incorporate the localized structure of robot interactions, show the most promise. We also present experiments that show that these two models can make effective use of larger warehouses operation datasets as the models are scaled up.

%% file: sections/01_Intro.tex
\section{Introduction}
\begin{figure}
\begin{subfigure}{.5\textwidth}
    \centering
    \includegraphics[height=4.5cm]{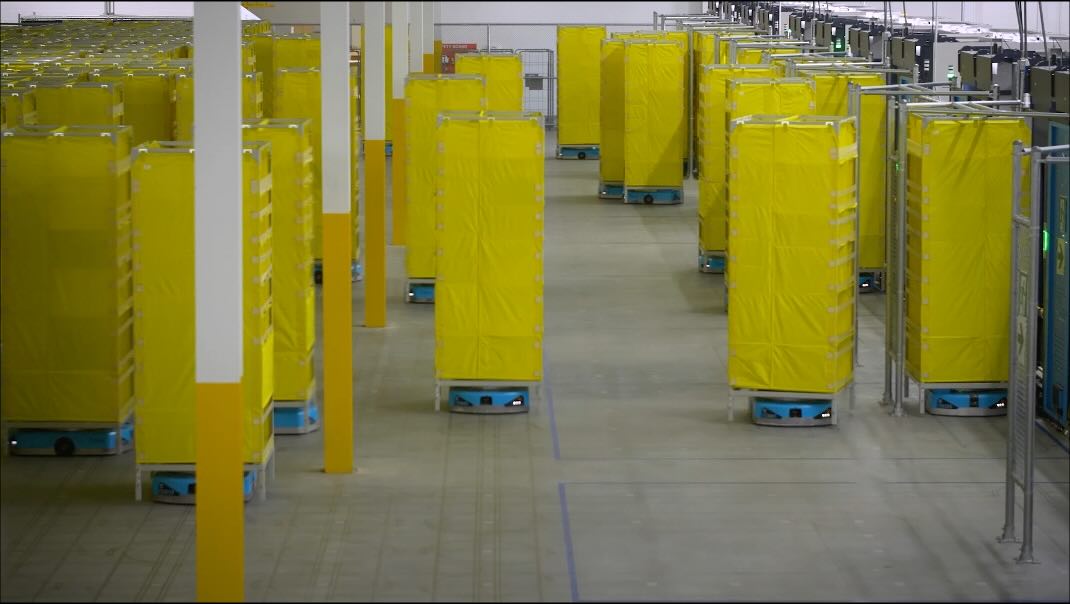}
    \label{fig:structured_floor_photos_storage}    
\end{subfigure}
\begin{subfigure}{.5\textwidth}
    \centering
    \includegraphics[height=4.5cm]{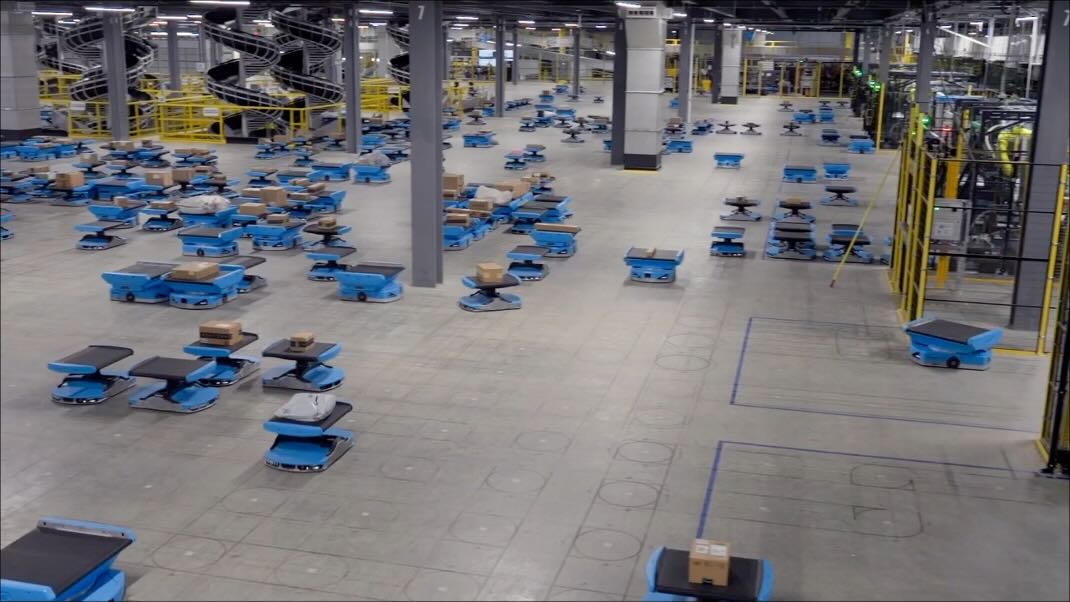}
    \label{fig:structured_floor_photos_sortation}    
\end{subfigure}
\caption{Robots carrying pods containing items on a storage floor (left) and packages on a sortation floor (right).}
\label{fig:structured_floor_photos}
\end{figure}

As of 2025, Amazon has deployed hundreds of thousands of robots throughout its global network of fulfillment and sortation warehouses (see Figure~\ref{fig:structured_floor_photos}). Each fulfillment warehouse floor must coordinate its fleet of mobile robots to transport shelves of inventory between storage locations and workstations to complete customer orders, while on sortation floors the robots pickup packages at workstations and drop them off at the right chute for their destination.
Effective coordination requires an understanding of complex multi-agent dynamics to enable proactive planning that avoids congestion and deadlocks, maximizes throughput, and completes customer orders on time. 

The discovery that scaling up deep neural networks and their training data can unlock unprecedented capabilities in natural language understanding ~\cite{brown2020language}, speech recognition ~\cite{zhang2022bigssl}, vision~\cite{radford2021learning}, and reasoning~\cite{openai2024o1} has recently led to efforts to apply the same data-driven scaling approach to robotic systems \cite{WaymoScaling2025, black2024pi_0,oneill2024open,ghosh2024octo}. Inspired by these efforts, we explore the design space of multi-agent foundation models to support the development of the next generation of robot fleet coordination algorithms. Just as foundation models pretrained on simple tasks such as masked text prediction can be adapted to applications via supervised fine-tuning, reinforcement learning, knowledge distillation, or in-context learning, our multi-agent fleet models are designed to serve as a foundation for many applications in our warehouses, such as congestion forecasting, adaptive routing, and proactive rescheduling. 

By using multi-agent forecasting as a pretraining objective, we expect these foundation models to learn how task assignments shape movement patterns, how congestion forms, and how local interactions propagate system-wide. With a large deployed fleet, multiple generations of mobile robots, hundreds of warehouses with diverse layouts and processes, and daily, weekly, and seasonal operational cycles, we expect our dataset to contain the kind of diversity that is necessary to provide a strong foundation for downstream tasks. 

This paper aims to address the following question: What are the most important architectural design choices for multi-agent foundation models, and how do they impact prediction accuracy and scaling? Specifically, we vary and study the impact of the extent of temporal and spatial context, the choice of uniform temporal snapshots or event-based updates, the choice of agent-centric or global state representations, the choice of state prediction or action prediction, and how information is propagated across time, space, and agents. 
The result is \DeepFleet, a suite of four model architectures, each embodying a distinctive inductive bias for learning how mobile robot fleets move in structured warehouse environments (see Table \ref{tab:design_matrix} in Section \ref{sec:system_overview}):

\vspace{-0.15cm}
\begin{itemize}
\item The \emph{robot-centric} (RC) model conditions a causal transformer on each robot's history of its own state, actions, and its local neighborhood to predict each robot's next action.

\vspace{0.15cm}
\item The \emph{robot-floor} (RF) model uses cross-attention between robot features and floor features to jointly leverage neighborhood information and global context.

\vspace{0.15cm}
\item The \emph{image-floor} (IF) model uses convolutional encoding over an image-like grid of the entire floor with separate channels for different robot and floor features.

\vspace{0.15cm}
\item Finally, the \emph{graph-floor} (GF) model uses graph neural network (GNN) and edge-conditioned  transformer layers to capture robot-to-robot and robot-to-floor relationships. 

\end{itemize}
\vspace{-0.15cm}

Our experimental results suggest that the robot-centric and graph-floor models, with limited spatial context, use their parameters more efficiently than the image-floor and robot-floor models, which provide full spatial context to each robot. They also show that convolutional features struggle to capture the dynamics of the floor, at least in a representation in which each robot is modeled as a single pixel. Initial scaling experiments for the robot-centric and graph-floor models suggest that there is room to scale up model size and dataset usage to improve model performance. We were able to extrapolate scaling curves from two orders of magnitude for the graph-floor model, while further experiments are needed to determine quantitative scaling laws for the robot-centric model.

The rest of the paper is organized as follows. Section~\ref{sec:related} describes related work. Section~\ref{sec:system_overview} provides a mathematical formulation of the multi-agent prediction problem. In Section~\ref{sec:results}, we describe our findings, including comparative evaluations and scaling experiments. In Sections~\ref{sec:rc}--\ref{sec:gf_model}, we describe each of the four architectures in detail. We offer concluding remarks in Section~\ref{sec:conclusion}.

%% file: sections/02_Related_Work.tex
\section{Related Work}
\label{sec:related}
\paragraph{Multi-Agent Trajectory Forecasting.} Significant work has been dedicated to modeling multi-agent interactions in trajectory forecasting, particularly in domains such as pedestrian motion prediction, autonomous driving, and social navigation. Early works such as Social-LSTM \cite{alahi2016social} and Social-GAN \cite{gupta2018social} captured inter-agent dynamics via recurrent neural networks with interaction-aware pooling. Transformer-based architectures \cite{giuliari2021transformer, salzmann2020trajectronpp}, variational models \cite{mangalam2021goals, tang2021collaborative}, and hybrid methods combining GNNs with probabilistic forecasting \cite{salzmann2020trajectron++, wang2024simple} have since improved generalization and multimodal prediction capabilities. However, these models are primarily applied to small-scale environments with limited agent counts and are not designed to scale to the thousands of agents encountered in warehouse scenarios.

\paragraph{Robotic Fleet Management in Warehouses.} 

Research in warehouse robotics has historically focused on task assignment, multi-agent pathfinding (MAPF) \cite{stern2019multi}, and traffic management. Until recently, the most successful techniques have been heuristic-based methods such as conflict-based search (CBS) \cite{boyarski2015icbs,li_eecbs_2021}, prioritized planning \cite{silver_cooperative_2021,ma_searching_2019}, MAPF-LNS \cite{li_anytime_2021}, priority inheritance with backtracking (PIBT) \cite{okumura_priority_2022}, LaCAM3 \cite{okumura_engineering_2024} and their variants — these often achieve strong performance with abstract robot models but struggle to scale on realistic robot models that account for dynamics and physical constraints \cite{yan_multi-agent_2025,zhang_efficient_2023}.

 Learning-based techniques  have also emerged to address path planning and task reallocation \cite{fung2019coordinating, choudhury2022dynamic,sartoretti_primal_2019,liu_mapper_2020,damani_primal_2021,wang_scrimp_2023,wang_mobile_2020,tang_railgun_2025} (see \cite{alkazzi_comprehensive_2024} for a review of learning-based MAPF through 2024), often combining imitation learning (IL) to learn behavior from heuristic MAPF solvers with reinforcement learning (RL) to learn improved behavior.

In 2024,  Yan \& Wu \cite{yan_neural_2023} and Veerapaneni \textit{et al.} \cite{veerapaneni_improving_2024} independently proposed using learned action prediction models to guide heuristic search techniques; meanwhile, Zhang \textit{et al.} proposed providing guidance with a model that generates graph edge weights \cite{zhang_guidance_2024}. These learned guidance mechanisms steer the search by replacing myopic heuristics with data-driven ones, leading to faster runtimes and better solutions. 

This combination of search and learning is emerging as a powerful tool in MAPF, propelling state of the art techniques such as SILLM \cite{jiang_deploying_2025}, SSIL \cite{veerapaneni_work_2024}, online GGO \cite{zang_online_2025}, and LaGAT \cite{jain_graph_2025}. Notably, LaGAT \cite{jain_graph_2025} demonstrates that even state-of-the-art learning-only methods like MAPF-GPT \cite{andreychuk2025mapf} can be integrated into search frameworks, though search efficiency varies with model complexity.

The unique scale of both our warehouses and the data we collect from them allows us to truly test the limits of learning-only methods; in Section \ref{sec:results} we describe scaling laws over several orders of magnitude of model and training dataset size.

\paragraph{Spatiotemporal Modeling and Representations.} Recent work in spatiotemporal modeling has produced powerful neural representations for structured environments. TrajectoryCNN \cite{liu2020trajectorycnn}, Scene-Transformer \cite{ngiam2021scene}, and ST-GNNs \cite{bui2022spatial} illustrate the utility of combining convolutional or graph-based spatial reasoning with temporal sequence modeling. In robotics, structured state-space models (SSMs) \cite{gu2023mamba, bhirangi2024hierarchical} and graph-based encodings \cite{dik2024graph} have improved data efficiency and temporal coherence.
Notably, GraphCast \cite{lam2023learning} and GenCast \cite{price2023gencast} have demonstrated the effectiveness of graph-based and generative models for large-scale spatiotemporal prediction, respectively excelling at deterministic and probabilistic forecasting in highly complex systems. 
Nonetheless, these methods often lack scalability, task generalization, or the integration of diverse spatial and temporal contexts required for warehouse-scale fleet modeling.

\paragraph{Scaling Laws and Compute-Optimal Models.} The rise of foundation models was enabled by empirical studies \cite{kaplan2020scaling} that demonstrated that language model performance followed predictable scaling laws with respect to compute, model size, and data. After further refinement \cite{hoffmann2022training}, these principles have been used to inform the scaling of large multi-modal models such as PaLM \cite{chowdhery2022palm} and LLaMA \cite{grattafiori_llama_2024}.

\paragraph{Foundation Models in Robotics:} Foundation models are also achieving state-of-the-art capabilities in robotics. Models such as RT-1 \cite{brohan2022rt}, RT-2 \cite{zitkovich2023rt}, VIMA \cite{jiang2022vima}, and PerAct \cite{shridhar2023perceiver} have demonstrated strong performance in manipulation and visuomotor policy learning. 
Flamingo \cite{alayrac2022flamingo} has enabled visual reasoning for embodied agents, while $\pi_0$ \cite{black2024pi_0}, Octo \cite{ghosh2024octo}, and Helix \cite{helix2025vision} have extended this foundation model approach toward multi-task, multi-embodiment settings. 
DeepMind's work on robotics foundation models and scalable policy architectures \cite{reed2022generalist} highlights the increasing effort to unify policy learning, perception, and reasoning in a single framework. 
These models emphasize generalization, multi-task capability, and real-world robustness---but are still primarily focused on manipulation or low-agent-count systems. Our work complements and extends these efforts by addressing fleet-scale modeling across thousands of agents in industrial settings.

%% file: sections/03_Problem_Formulation.tex
\section{Problem Formulation}
\label{sec:system_overview}

We consider a generalized warehouse environment populated by a large fleet of mobile robots.  
The layout is modeled as a directed graph \(G = (V, E)\), where each vertex \(v \in V\) represents one of \(M\) defined locations on the floor, and each directed edge \((u,v) \in E\) denotes an admissible motion between these locations. The fleet with $N$ mobile robots (\(N \ll M\)) traverses this graph to execute logistics tasks issued by a centralized coordinator, 
which involve picking up an object from one vertex, carrying it through the graph, and eventually unloading the object at another vertex. 

\begin{figure}[t]
\centering
\includegraphics[width=0.99\linewidth]{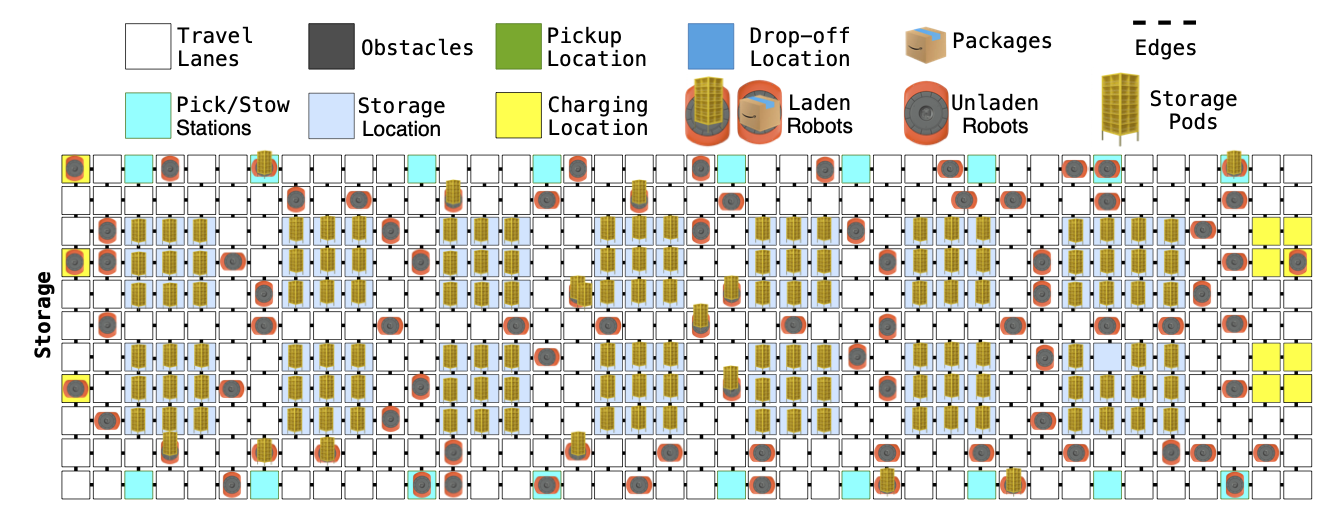}

\vspace{.15in}

\includegraphics[width=0.99\linewidth]{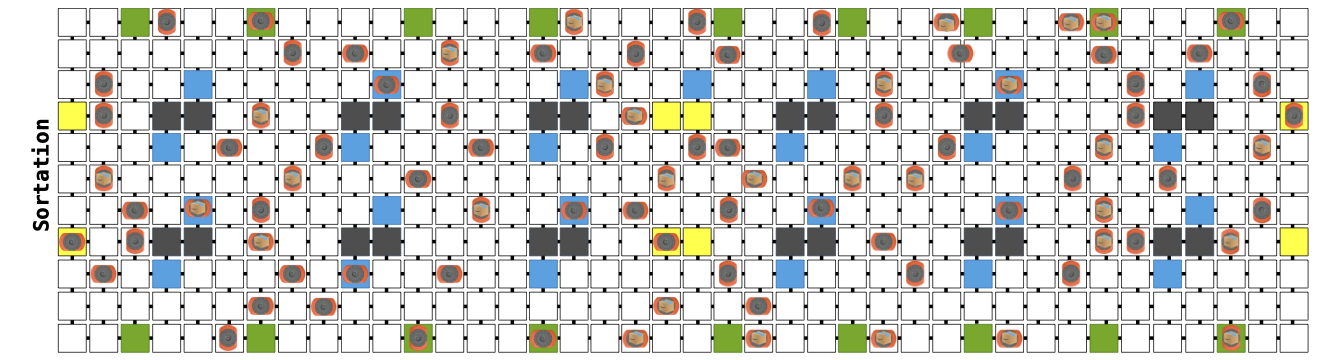}
\caption{Scaled-down examples of storage (top) and sortation floors (bottom). Robots navigate a spatial field of discrete locations, each with their own attributes.}
\label{fig:structured_floor}
\end{figure}

On our storage floors, objects are pods containing inventory (see Figure~\ref{fig:structured_floor}, top). Loading a pod requires a robot to dock under it and actuate a lifting mechanism. Pods are stored in designated areas on the floor, and robots transport them between storage and workstations where items are picked out of the pod or stowed into it.
On sortation floors, robots bring packages from pickup stations to drop-off locations (see Figure~\ref{fig:structured_floor}, bottom). These robots carry one package at a time and are equipped with a conveyor belt, which they use to eject the package at the drop-off location.
Storage and sortation robots have similar footprints and dynamics, but the emerging dynamics of the two types of floors differ: movement on storage floors tends to be more spatially constrained by pod storage areas, while congestion on sortation floors can emerge from spikes in demand for certain drop-off locations.

\DeepFleet models are trained using production data from both types of floors, using the following representation to unify both storage and sortation floors.

\paragraph{Robot state.}
At time \(t\), each robot \(i\) is described by a state vector
\(
\mathbf{s}_t^i = (p_t^i, \theta_t^i, g_t^i, \ell_t^i)
\),
where \(p_t^i\in V\) is its current position, \(\theta_t^i\) its heading, \(g_t^i\) its goal location, and \(\ell_t^i\) is its load status (whether the robot is carrying a pod/package). 

\paragraph{Vertex features.} Vertex features include the vertex location on the floor in a global reference frame, and whether the vertex is used for travel, charging, or other tasks. For storage floors, a vertex may be designated for pod storage or for picking/stowing from a pod. On sortation floors, vertices can be pickup or drop-off locations.

\paragraph{Floor state.} 
Aggregating over all vertices yields a \emph{floor state}
\(
S_t \in \mathbb{R}^{M \times d},
\)
where each row encodes static vertex features and any dynamic robot features associated with that vertex (with feature dimension \(d\)).  
This unified notation accommodates both \emph{floor-centric} views (a dense vertex grid) and \emph{robot-centric} views (sparse robot states mapped onto vertices).

\paragraph{Fleet dynamics.}
Robot motions are discrete as they transitions from vertex \(p_t^i\) to an adjacent vertex along a directed edge in \(E\), but they are not assumed to be synchronous.  
Although the graph \(G\) imposes static constraints, real-time fleet behavior is shaped by dynamic factors such as task arrivals, other robots' trajectories, and occasional obstacles. Coupling between hundreds of agents produces emergent phenomena (congestion, deadlocks, traffic waves) that delay robot missions if not anticipated.

\paragraph{Forecasting objective.}
Let \(K\) denote the historical window length and \(H\) the prediction horizon.  Over this window, we observe a sequence of floor states $S_{t-K:t} = (S_{t-K},\ldots,S_t)$ and actions $A_{t-K:t-1} = (A_{t-K}, \ldots, A_{t-1})$, where for any time step $t$ the set of robot actions $A_t$ is used to transition from state $S_t$ to $S_{t+1}$. For a robot $i$, $a_t^i \in A_t$ is one of:
\begin{itemize}[leftmargin=*,nosep,itemsep=2pt]
    \item move forward (in the direction of $\theta^{i}$) by $k$ vertices; 
    \item rotate by -90, +90, or 180 degrees;
    \item load or unload an object; or
    \item wait at the current vertex.
\end{itemize}
We aim to learn a function $F_\theta$ that predicts the states and/or actions over the next $H$ steps:
\begin{equation}
\bigl( \, \hat{A}_{t:t+H-1},\; \hat{S}_{t+1:t+H} \,\bigr) = F_\theta\bigl(A_{t-K:t-1}, S_{t-K:t}\bigr)
\end{equation}
Some such functions can be factorized into an \textit{autoregressive} form, meaning that only a time step is predicted, and then the prediction is fed back to condition the next time step's prediction. Because states and actions are tightly coupled, different model architectures vary in the precise form of the inputs. For example, a model could operate only on the states as input and output, so 
\(
\hat{S}_{t+1:t+H}, 
\;=\;
F_\theta\bigl(S_{t-K:t}\bigr).
\)

On the other hand, a model could use states and actions to predict the next action, then use a deterministic \textit{environment model} E to evolve the state:
\begin{equation*}
\left\{
\begin{alignedat}{1}
\hat{A}_{t} & = F_\theta\bigl(S_{t-K:t}, A_{t-K:t-1}\bigr) \\
\hat{S}_{t+1} & = E(S_t, \hat{A}_{t}).
\end{alignedat}\right.
\end{equation*}
This approach is used by both the RC and GF models.

The unifying framework for our \DeepFleet models is the task of predicting future states and actions from previous states and actions. To do this effectively across our diverse large-scale dataset requires a model to learn rich representations, which we eventually hope to use for applications beyond simulation. 

Table \ref{tab:design_matrix} shows how our four distinct model architectures use different approaches to spatial and temporal representation to approach the problem with different inductive biases. The robot-centric and graph-floor models use asynchronous event-based movement data while the image-floor and robot-floor model use snapshots sampled uniformly in time. The robot-centric model uses a local neighborhood view of each robot and predictions actions, while the image-floor model uses a whole-floor view and predicts states; the other two models use a combination of local and global views and predict states and actions jointly.

Further details of each model architecture are in Sections \ref{sec:rc} (robot-centric), \ref{sec:rf} (robot-floor), \ref{sec:if_model} (image-floor), and \ref{sec:gf_model} (graph-floor). First, we summarize our findings.

\begin{table}[htbp]
\centering
\caption{Comparison of Model Architectures for Robotic Applications}
\resizebox{\textwidth}{!}
{
\begin{tabular}{x{10em}x{10em}x{6em}x{6em}x{10em}x{6em}x{6em}x{9em}x{4em}x{4em}}
\toprule
\textbf{Model Architecture} & \textbf{Temporal Modeling} & \multicolumn{2}{c}{\textbf{Temporal Approach}} & \textbf{Spatial Modeling} & \multicolumn{2}{c}{\textbf{Spatial Scope}} & \textbf{Prediction Target} & \multicolumn{2}{c}{\textbf{Output Type}} \\
 & & \textit{Fixed-Time Snapshots} & \textit{Event-Based Updates} & & \textit{Local Agent Centric} & \textit{Global Floor Level} & & \textit{Future States} & \textit{Future Actions} \\
\midrule
Robot-Centric (Decision Transformer) & Sequential causal modeling & \xmark & \cmark & Individual robot perspective & \cmark & \xmark & Decision-making focus & \xmark & \cmark \\
\midrule
Image-Floor (CNN+GPT) & Grid-based temporal evolution & \cmark & \xmark & Spatial convolutions over floor & \xmark & \cmark & Floor state evolution & \cmark & \xmark \\
\midrule
Robot-Floor (Multimodal GPT) & Fixed snapshots with transformer atten. & \cmark & \xmark & Hybrid: combines both perspectives & \cmark & \cmark & Multiple output heads for both & \cmark & \cmark \\
\midrule
Graph-Floor (GNN + Transformer) & Graph-based message passing & \xmark & \cmark & Relational graph structure & \cmark & \cmark & Graph node predictions & \cmark & \cmark\\
\bottomrule
\end{tabular}
}
\label{tab:design_matrix}
\end{table}

%% file: sections/04_Findings.tex
\section{Findings}
\label{sec:results}
In this section, we provide an overview of our findings, before continuing with detailed descriptions of the four model architectures.
We evaluated the following instantiations of our model architectures:
\begin{itemize}[leftmargin=*,nosep,itemsep=2pt]
    \item A robot-centric (RC) model with 97M parameters using a spatial context of 30 nearest robots, 100 nearest markers, 100 nearest objects, a temporal context of 5 state/action pairs, trained on about 5 million robot-hours of data.
    \item A robot-floor (RF) model with 840M parameters using 10 state/action pairs per robot and one snapshot of whole-floor context trained on about 700,000 robot-hours of data.
    \item An image-floor model (IF) model with 900M parameters using 60 seconds of whole-floor context trained on about 3 million robot-hours of data.
    \item A graph-floor (GF) model with 13M parameters using 4 seconds of whole-floor context trained on about 2 million robot-hours of data.
\end{itemize}

\begin{table}
\centering
\caption{Model performance comparison. Lower values indicate better performance across all metrics.}
\label{tab:comp_results}
\begin{tabular}{cccccc}
\toprule
\textbf{Model} & \textbf{Parameter count} & \multicolumn{3}{c}{\textbf{DTW Deviation}} &  \textbf{CDE (\%)} \\
 & & Position & State & Timing &  \\
\midrule
Robot-Centric (RC) & 97M & \textbf{8.68} & \textbf{0.11} & 14.91 &  \textbf{3.40} \\
Robot-Floor (RF) & 840M & 16.11 & 0.23 & \textbf{6.53} &  9.60 \\
Image-Floor (IF) & 900M & 25.02  & 1.58 & 48.29 &  186.56 \\
Graph-Floor (GF) & 13M & 10.75 & 0.75 & 21.35 &  14.22 \\
\bottomrule
\end{tabular}%
\end{table}

We evaluated these models with metrics that quantify the degree to which the behavior predicted by the models matches the behavior in our test dataset, which consisted of 7 days across 7 warehouse floors that were held out during training. We used model inference to iteratively roll out robot trajectories 60 seconds into the future for each sample in the test partition of the dataset, and then compared them to the ground truth trajectories using the following metrics:
\begin{itemize}[leftmargin=*,nosep,itemsep=2pt]
    \item {\bf Dynamic time warping (DTW)} distance \cite{muller2007dynamic} across multiple dimensions: robot position and speed, state, and timing of load and unload events. Its units are the same as the underlying measurement, e.g., the DTW error for position is in meters, and measures the average distance between the true and estimated trajectory after the optimal temporal alignment is found.
    \item  \textbf{Congestion delay error (CDE)}, or the relative error between the proportion of time robots are delayed by others from an inference rollout and the ground truth.  These delays are calculated as
    $(t_{\mathrm{total}} - t_{\mathrm{free\_flow}})/t_{\mathrm{total}}$, where $t_{\mathrm{total}}$ is the actual travel time of a set of robot trajectories and $t_{\mathrm{free\_flow}}$ is the counterfactual travel time if robots could occupy the same space and not interfere with the motion of others.
\end{itemize}

DTW measures the ability of the models to predict the motion and state of the robots, which was the pretraining objective for all of the models. Congestion delay is a stand-in for the class of operational metrics we want the model to be able to predict and eventually optimize; CDE measures the error in a model's ability to generate floor dynamics that have the right amount of congestion.

Table \ref{tab:comp_results} presents the results. Overall, the RC model achieved the best performance across the most metrics. The RF model was able to achieve strong performance on state and timing error, though with a higher parameter count. Meanwhile, the GF model is fairly competitive despite its low parameter count. The IF model struggled to accurately model the dynamics, and qualitatively the rollouts included large jumps in robot positions from one second to the next. 

The models were sized to fit on the available hardware during this phase of the project, and the varying architectures caused big differences in the model sizes that makes direct comparison difficult. However, we can still draw some conclusions from these results. First, we now believe that the image-based approach that treats each location as a pixel and uses convolutional features may not provide the right inductive bias to model robot fleet interactions. Also, both the IF and RF results suggest that providing complete spatial context to every robot is an inefficient use of parameters. Instead, the RC and GF models allow local interactions that can be propagated to gain global understanding, and these models were able to achieve superior performance at a much smaller size.

\begin{figure}
\begin{subfigure}{.5\textwidth}
    \centering
    \includegraphics[height=8.6cm]{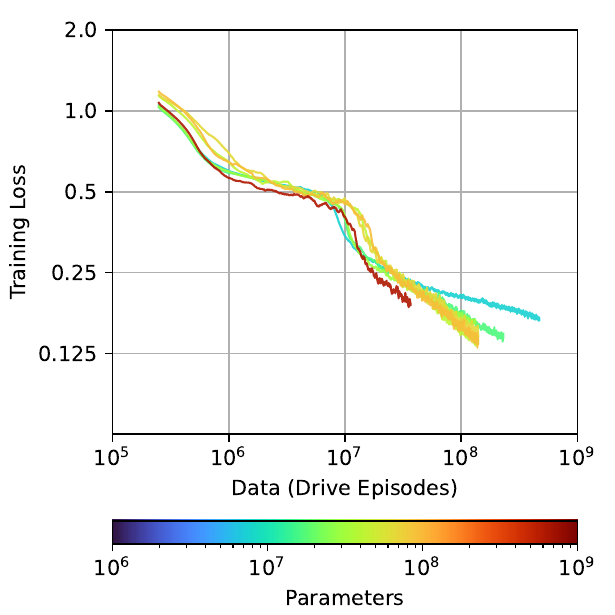}
    \caption{Robot-centric}
    \label{fig:loss_vs_data_RC}    
\end{subfigure}
\begin{subfigure}{.5\textwidth}
    \centering
    \includegraphics[height=8.6cm]{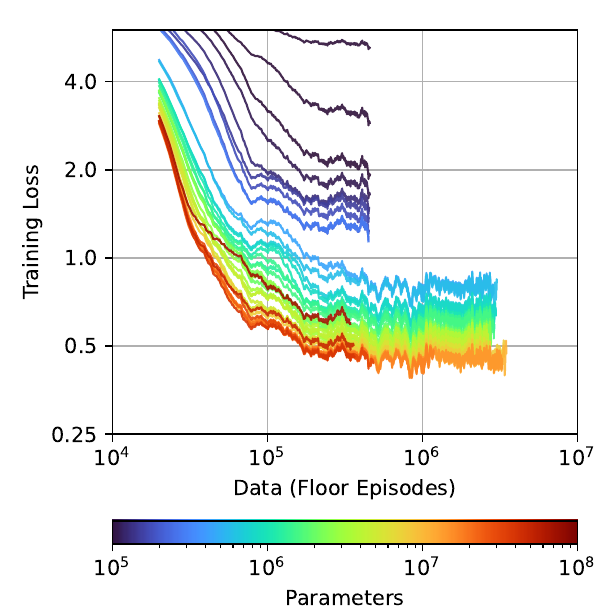}
    \caption{Graph-floor}
    \label{fig:loss_vs_data_GF}    
\end{subfigure}
\caption{Training loss for the robot-centric model (a)  and the graph-floor model (b) as a function of data size, showing that larger models are effectively using the available data.}
\label{fig:loss_vs_data}
\end{figure}
%
\begin{figure}
\begin{subfigure}{.5\textwidth}
    \centering
    \includegraphics[height=8.6cm]{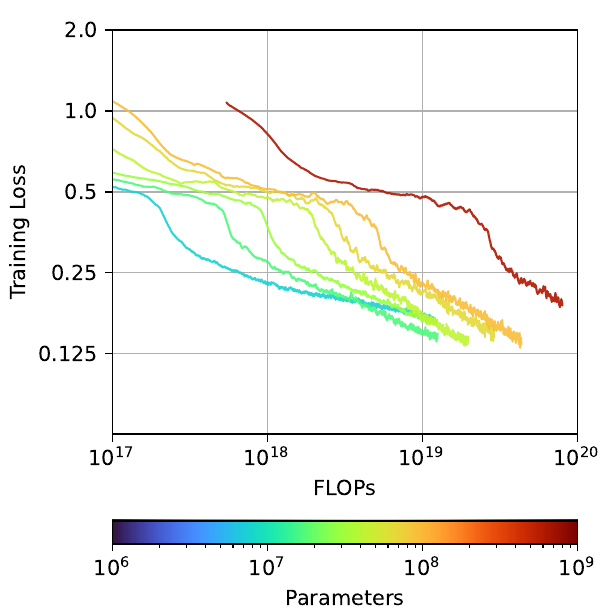}
    \caption{Robot-centric}
    \label{fig:loss_vs_flop_RC}    
\end{subfigure}
\begin{subfigure}{.5\textwidth}
    \centering
    \includegraphics[height=8.6cm]{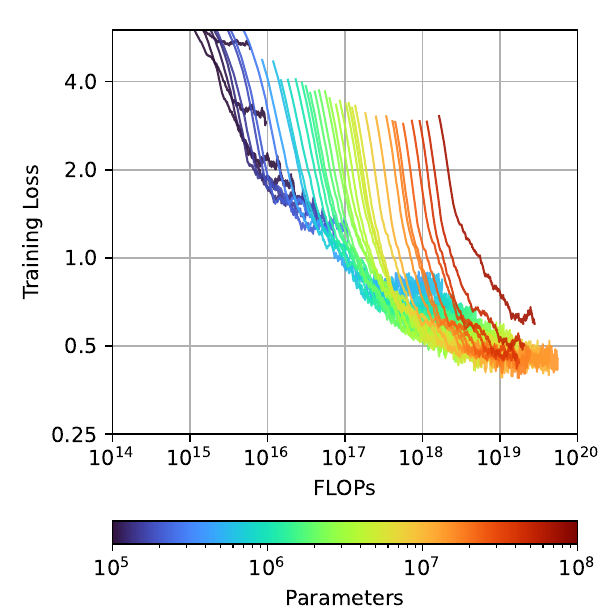}
    \caption{Graph-floor}
    \label{fig:loss_vs_flop_GF}    
\end{subfigure}
\caption{Training loss for the robot-centric model (a) and the graph-floor model (b) as a function of training FLOPS.}
\label{fig:loss_vs_flop}
\end{figure}
%
\begin{figure}
\begin{subfigure}{.5\textwidth}
    \centering
    \includegraphics[width=8.5cm]{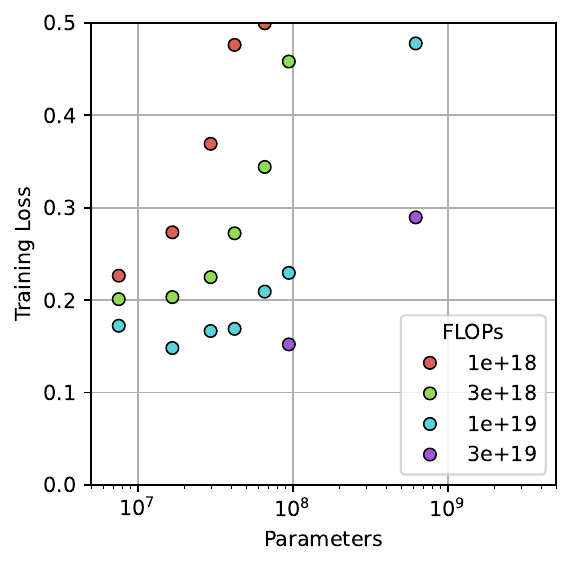}
    \caption{Robot-centric}
    \label{fig:isoflop_RC}    
\end{subfigure}
\begin{subfigure}{.5\textwidth}
    \centering
   \includegraphics[width=8.5cm]{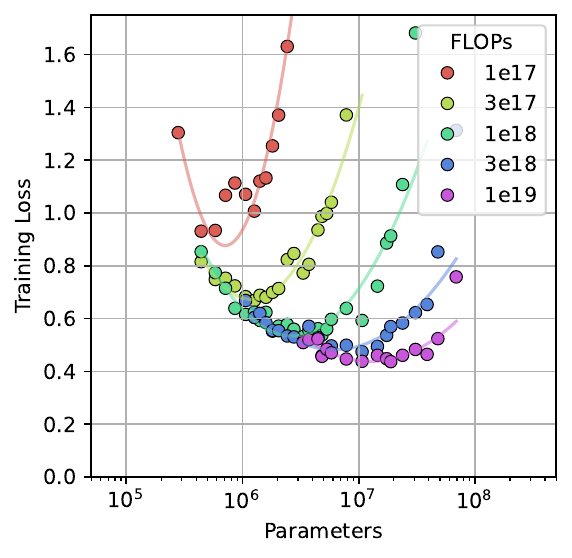}
   \caption{Graph-floor}
    \label{fig:isoflop_GF}    
\end{subfigure}
\caption{IsoFLOP curves for the robot-centric model (a) and the graph-floor model (b). While the robot-centric model needs longer training runs to see full parabolic curves, the graph-floor model allows for extrapolation from two orders of magnitude.}
\label{fig:isoflop}
\end{figure}
%
\begin{figure}
\begin{subfigure}{.5\textwidth}
    \centering
    \includegraphics[width=8.5cm]{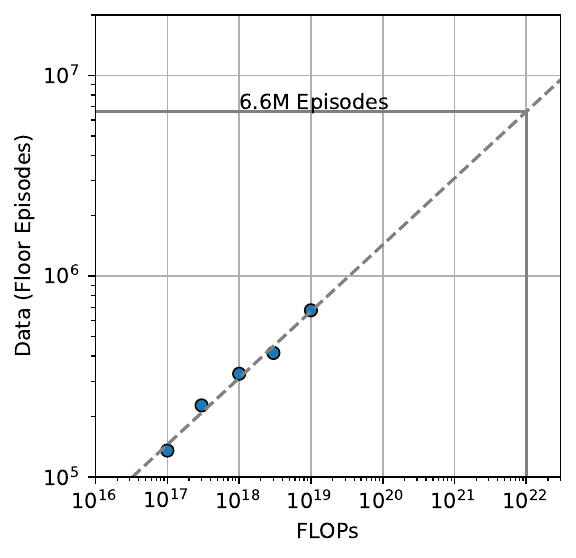}
    \label{fig:data_vs_flop_GF}    
\end{subfigure}
\begin{subfigure}{.5\textwidth}
    \centering
   \includegraphics[width=8.5cm]{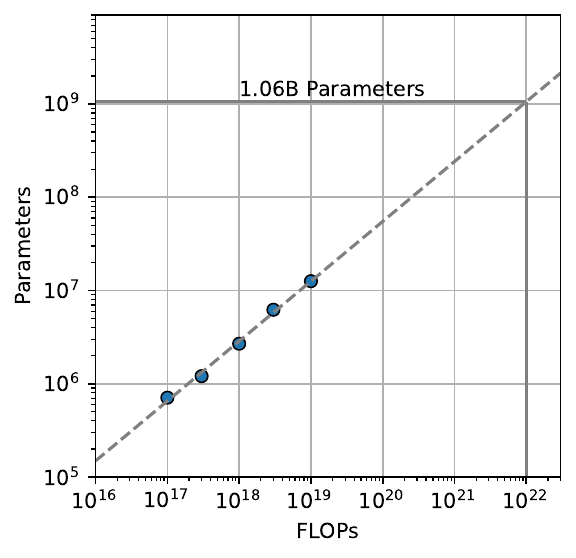}
    \label{fig:parameter_vs_flop_GF}    
\end{subfigure}
\caption{Extrapolation of optimal dataset size and model size based on isoFLOP curves for the graph-floor model. The power-law fit on each curve suggest that, for example, when scaling to a 1.06B model, we should train with approximately 6.6M floor episodes in order to make the best use of training compute; this will correspond to a training run of $10^{22}$ FLOPs, which is easily attainable in a few weeks on a small cluster.}
\label{fig:optimal_flops_vs_parameters}
\end{figure}

Based on these results, we conducted scaling experiments for the RC and GF architectures to extrapolate how more data and larger models using these architectures will improve performance. These experiments are modeled after scaling experiments in language modeling \cite{kaplan2020scaling,hoffmann2022training}, which have been used \cite{grattafiori_llama_2024} to scale to extremely large models and datasets using architectural decisions made from smaller models.

The loss curves with respect to data in Figure \ref{fig:loss_vs_data} show that the larger models we tested achieve better loss values given equal data, suggesting that the architectures are improving in performance with scale. The loss curves with respect to compute in Figure \ref{fig:loss_vs_flop} paint a slightly different picture for the robot-centric and graph-floor models. The robot-centric models appear to need further training with more data to show the full benefits of more scaling; the graph-floor models have been trained on enough data to show a clear loss envelope that can be used to extrapolate to larger models and dataset sizes.

The isoFLOP curves for the robot-centric and graph-centric models, shown in Figure \ref{fig:isoflop}, illustrate this. For the robot-centric model, we need to train longer to complete the parabolic curves and estimate optimal model and dataset sizes at each FLOP level; for this reason we did not do a fit or extrapolation on these curves. 

We did, however, fit parabolas to the isoFLOP curves for the graph-centric model (Figure \ref{fig:isoflop_GF}). This allowed us to estimate the optimal model size (and thence the dataset size) at each FLOP level. The results are shown in Figure \ref{fig:optimal_flops_vs_parameters}. The optimal dataset and model size as a function of training FLOPs follows a power law for two orders of magnitude. Extrapolating this power law suggests that the optimal use of $10^{22}$ FLOPs would be to train a 1B model on approximately 6.6 million floor episodes. This could be achieved with a few weeks of training time on a small cluster of advanced GPUs.

We plan to use these findings for the further development of the graph-floor model, and do further scaling experiments with the robot-centric model to extrapolate its scaling curve.

%% file: sections/05_Robot_Centric_Model.tex
\section{Robot-centric Model}
\label{sec:rc}

The robot-centric (RC) model occupies the \emph{event-based, asynchronous} corner of the design matrix. It uses an agent-centric, ego-frame neighborhood spatial view to predict each robot's next action, and is paired with a deterministic environment model to propagate the state.

At each time step $t$ a robot $i$'s observations include a set $\mathcal{R}_t^i$ of its $K_r$ nearest neighbor robots, $\mathcal{P}_t^i$ of its $K_p$ nearest neighbor objects, and $\mathcal{X}_t^i$ of its $K_x$ nearest neighbor vertices. The state at each time step $t$ is represented by a tuple of object embeddings $\mathbf{o}_t^i$ associated with robot $i$ and its neighborhood:
\begin{equation}
\mathbf{o}_t^i = \left(\mathbf{r}_t^i, (\mathbf{r}_t^j)_{j \in \mathcal{R}_t^i}, (\mathbf{p}_t^k)_{k \in \mathcal{P}_t^i}, (\mathbf{x}_t^{\ell})_{\ell \in \mathcal{X}_t^i}\right)
\end{equation}
where $\mathbf{r}_t^i$ is the embedding vector of the ego robot's state $\mathbf{s}_t^i$,  $\mathbf{r}_t^j$ are the embeddings of neighborhood robots' states $\mathbf{s}_t^j$, $\mathbf{p}_t^j$ are the the locations of neighborhood objects, and $\mathbf{x}_t^j$ are neighborhood vertex embeddings. Here we normalize these states to induce translation- and rotation-invariance. (Note that the time steps $t$ are discrete indices which correspond to sequential, irregular updates from robot $i$; for the dynamic object and robot neighborhoods, the nearest neighbors are determined by their last state before robot $i$'s time step $t$).

This object-centric representation allows for easy extension---new types of objects or attributes can be introduced by expanding the feature vectors or adding additional token types. Each object token is then embedded into a shared latent space via a learnable embedding function. Since all robots of the same type are physically and behaviorally equivalent, a single model can be used to predict their behavior across the dataset. 

\subsection{Model Framework}
The RC model is composed of two key components (not counting the environment model, which is described below). 
The encoder $\mathcal{E}_\theta$ uses a standard transformer architecture to map the tuple of neighborhood embeddings $\mathbf{o}_t^i$ into a single latent embedding:
\begin{equation}
\mathbf{h}_t^i = \mathcal{E}_\theta\left(\mathbf{o}_t^i\right)
\end{equation}
The decoder $\mathcal{D}_\theta$ then takes an alternating sequence of hidden state embeddings and action embeddings and autoregressively predicts the next action:
\begin{equation}
\hat{\mathbf{a}}_{t}^i = \mathcal{D}_\theta(\mathbf{h}_{t-K}^i, \mathbf{a}_{t-K}^i, \ldots \mathbf{h}_{t-1}^i, \mathbf{a}_{t-1}^i, \mathbf{h}_t^i)
\end{equation}
This is similar to the decision transformer approach~\cite{chen2021decision}, but since we are in a behavior cloning setting our sequence alternates between state and action embeddings without returns-to-go. Note that since the action vocabulary is discrete, and the model outputs a probability distribution over the actions.

\subsection{Training Objective}
The model is trained using behavior cloning with teacher forcing. The objective is to minimize the negative log-likelihood of observed actions under the model:
\begin{equation}
\mathcal{L}_{RC} = - \sum_{i=1}^{N} \sum_{t} \log p_\theta\left(\mathbf{\hat{a}}_t^i = \mathbf{a}_t^i \Big| \mathbf{o}_{t-K:t}^i, \mathbf{a}_{t-K:t-1}^i\right)
\end{equation}
This encourages the model to replicate observed behavior over the training window.

\begin{figure}
\centering
\includegraphics[width=0.99\linewidth]{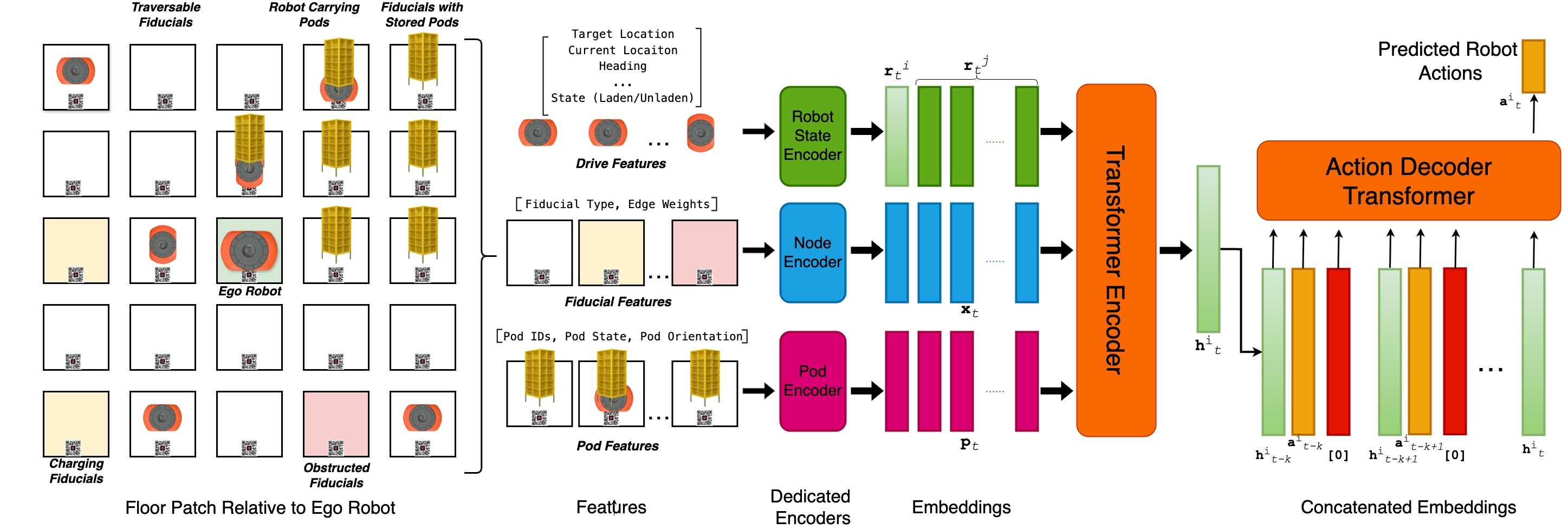}
\caption{
Robot-Centric model architecture. The model utilizes a Transformer Encoder to build the latent space for the robot's localized state, and robot tokens are passed to a Decoder Transformer to generate actions autoregressively.}
\label{fig:rc_architecture}
\end{figure}

The RC model provides a compact and modular formulation for predicting robot actions from local observations. By decoupling the environment into structured object tokens and conditioning on ego-centric views, the model supports generalization across different robot instances, floor geometries, and task types. Its translation- and rotation-invariant representation reduces sample complexity and facilitates large-scale training with shared parameters across robot units. Moreover, the autoregressive formulation enables multi-step rollout and simulation of agent behavior for long-horizon forecasting.

\subsection{Inference Rollout}
At inference time the same RC model is applied in parallel to every robot on the floor. Given the current floor state $S_t$, we construct independent neighborhood windows $\mathbf{o}_{t-K:t}^1,\dots,\mathbf{o}_{t-K:t}^{N}$ and obtain action distributions $\hat a_t^1,\dots,\hat a_t^{N}$ in a single batched forward pass. Since this model predicts actions only, an environment model is needed to evolve the state. We use a deterministic environment that operates on the robots sequentially. For each time step, each robot attempts to apply its action, which may require reserving a set of vertices to move through. If any of those vertices is already reserved, the robot executes a wait action instead; otherwise it reserves the necessary vertices and updates its position. Aggregating all accepted state transitions yields the new floor state $S_{t+1}$, which then seeds the subsequent prediction step.  Because weights are shared across all $N$ agents, the overall computation scales linearly with fleet size.

%% file: sections/06_Robot_Floor_Model.tex
\section{Robot-floor Model}\label{sec:rf}
\begin{figure}[t]
\begin{subfigure}{\textwidth}
\centering
\includegraphics[width=0.99\linewidth]{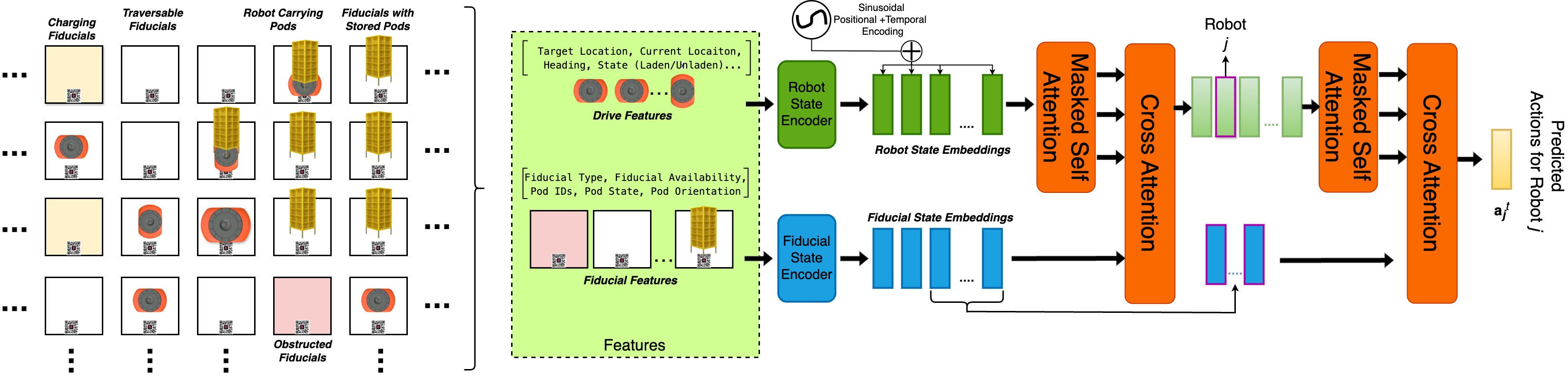}
\label{fig:rf_decoder}
\end{subfigure}
\caption{Robot-Floor model architecture. The model accepts an input sequence of robot tokens that self-attend to each other and cross-attend to a sequence of fiducial tokens before decoding the next action at the next time step. This figure depicts the decoder-only variant, which we explain in section \ref{sec-rf-model-framework}.}
\label{fig:rf_encoder_decoder}
\end{figure}

The robot-floor (RF) architecture (Figure~\ref{fig:rf_encoder_decoder}) occupies the \emph{fixed-time, synchronous} corner of the design matrix. At each time step $t$, system state is factorized into two heterogeneous token sets consisting of $N$ \emph{robot tokens} \(\mathbf r_{i,t} \) and $M$ contextual \emph{floor tokens} \(\mathbf x_{j,t} \) that correspond to a semi-static, physical floor elements that do not require prediction, but do impact fleet movement, such as vertex type, graph edges, objects and their locations. The model decodes each robot token \(\mathbf r_{i,t} \) into an action token \(\mathbf a_{i,t} \) after conditioning on robot-to-robot context and floor-to-robot context, via self-attention and cross-attention layers, respectively.

Each robot feature $\mathbf r_i$ concatenates absolute pose, target location, state (laden/unladen), and robot type identifiers. Each context feature concatenates absolute position, object properties, vertex type, and graph edge features. Raw features are then embedded via learnable projections $\phi:\mathbb R^{d_r}\!\!\rightarrow\!\mathbb R^{d}$ and $\psi:\mathbb R^{d_x}\!\!\rightarrow\!\mathbb R^{d}$, after which they share a common latent dimension $d$. 

Isolating the \(N\)-token \emph{robot stream} \(\mathbf R_t\) from the much longer \(M\)-token \emph{context stream} \(\mathbf X_t\) helps reduce quadratic self-attention cost from \(O(M^{2})\) to \(O(N^{2})\) (two orders of magnitude on a typical floor). Robots still access the full floor snapshot through a single cross-attention pass of cost \(O(N M)\), which also injects rich relational features such as edge distance or path cost~\cite{shaw2018,press2021alibi}. This enables global context where all robots can, in principle, attend to every object on the warehouse floor (inventory, other robots, vertices and their connectivity), enabling long-range contextual understanding of fleet behavior. The context encoder can append new modalities (e.g., no-go zones) to \(\mathbf X_t\) without touching the robot decoder, and the same model generalizes to arbitrary robot counts or floor sizes simply by changing \(N\) and \(M\).

\subsection{Model Framework} \label{sec-rf-model-framework}
The RF model treats next-action prediction as a sequence transduction task, mapping an input sequence of robot states to an output sequence of actions: \((\mathbf{r}_1, \ldots, \mathbf{r}_N) \mapsto (\mathbf{a}_1, \ldots, \mathbf{a}_N)\). To learn this mapping, interactions among robot and context tokens are modeled by two attention mechanisms. Self-attention among all robot tokens allows the model to learn \textit{robot-contextualized} representations $p(\mathbf{r}_i' | \mathbf{r}_1, \ldots, \mathbf{r}_N)$ that are predictive of next action (i.e., ``how robots influence each other's motion''). Cross-attention allows robot tokens to attend to the floor context around them to produce \textit{floor-contextualized} representations $p(\mathbf{r}_i' | \mathbf{x}_1, \ldots, \mathbf{x}_M)$ that are predictive of next action (i.e., ``how the structure of the floor environment conditions next action probabilities''). The use of cross-attention allows very large contextual attention independent of the decoded sequence length~\cite{jaegle2021perceivergeneralperceptioniterative}. 

Given a historical window of snapshots \(\mathbf R_{t-K:t},\mathbf X_{t-K:t}\) we have the option to explore two orthogonal attention axes: along the robot index or along the time index. Attention along the robot index interprets the input as a sequence of all $N$ robots at a single time step $t$. The input sequence is then self-attended among all robots and decoding of all $N$ actions for the next time step happens in parallel. On the other hand, attention along the time index interprets the input as a sequence of $K$ time steps for a single robot. The input sequence is then causally self-attended among all previous states and decoding the next action happens serially. For this work, we implemented the latter time-indexed method, which we call a \emph{decoder-only} variant.

\vspace{0.2em}

\vspace{0.2em}
\noindent\textbf{Decoder-only (per-robot temporal modeling).}
Let \( \mathbf z_{i,t-K:t}=[\mathbf r_{i,t-K};\dots;\mathbf r_{i,t}] \in\mathbb R^{K\times d} \)  be robot \(i\)'s local state history, to which we add a fixed sinusoidal position encoding. After projection with the embedding map \(\phi(\cdot)\), the sequence is processed by \(L\) stacked decoder blocks. In block \(l\) we write \(\mathbf h^{(l-1)}_{i,*}\) for the input sequence and obtain
\[
\mathbf h^{(l)}_{i,*}
  =\mathcal F_{\theta^{(l)}_{\text{PX}}}\!
     \Bigg(
        \mathcal F_{\theta^{(l)}_{\text{CX}}}\!
           \bigg(\,
              \mathcal F_{\theta^{(l)}_{\text{SA}}}
                 \Big(\mathbf h^{(l-1)}_{i,*}\Big)
              ,\,
              \psi(\mathbf X_t)
            \bigg),
        \phi(\mathbf R_t)
     \Bigg),
\]
where  

\medskip
\hspace*{1.2em}\(\mathcal F_{\theta^{(l)}_{\text{SA}}}\) is the causal self-attention over the history tokens;  

\hspace*{1.2em}\(\mathcal F_{\theta^{(l)}_{\text{CX}}}\) is the cross-attention to the frozen floor context
\(\psi(\mathbf X_t)\);  

\hspace*{1.2em}\(\mathcal F_{\theta^{(l)}_{\text{PX}}}\) is the cross-attention to the peer snapshot \(\phi(\mathbf R_t)\).

\medskip\noindent
After the final layer, the embedding of the last token \(\mathbf h_{i,t}\!=\!\mathbf h^{(L)}_{i,K}\) summarizes robot \(i\)'s past and current surroundings.  An output head shared across all robots, \(\mathcal F_{\theta_{\text{out}}}\), converts the embeddings to probabilities over the action vocabulary:
\[
\mathbf{\hat a}_{i, t}
  =\mathcal F_{\theta_{\text{out}}}\!\bigl(\mathbf h_{i,t}\bigr)
  \in\mathbb R^{|\mathcal A|}.
\]

\subsection{Training Objective}
The RF model is trained using cross-entropy loss between predicted and ground truth actions of individual robots:
\begin{equation}
\mathcal{L}_{\text{action}} = -\sum_{i=1}^{N} \sum_{j \in \mathcal{A}} \mathbf{a}^j_{i} \log(\hat{\mathbf{a}}^j_{i})
\end{equation}
where $\mathcal{A}$ is the set of possible robot actions (see Section~\ref{sec:system_overview}), $\mathbf{a}_{i}$ is the the ground truth one-hot encoded action for robot $i$, the superscript is used to index elements of a vector, and we elide the time subscript for simplicity.
The loss is differentiable w.r.t.\ the shared parameters \(\theta=\{\theta_{\text{SA}},\theta_{\text{CA}},\theta_{\text{CX}},
      \theta_{\text{PX}},\theta_{\text{out}}\}\), and gradients are accumulated across all robots and timesteps in the mini-batch.  
The rich floor context and peer interactions are learned end-to-end through this single action prediction signal.

\subsection{Inference Roll-out}
\label{sec:rf_inference}

At inference time, we decode one robot at a time. For each \(i\) we feed its state history
\(\mathbf z_{i,t-K:t}\) together with the frozen context
\(\bigl(\mathbf R_t,\mathbf X_t\bigr)\) through the decoder stack to produce \(\hat a_t^{\,i}\).  The procedure repeats asynchronously as fresh state is appended, and a cache of key-value memories enables \(O(1)\) time per step. 

In summary, the RF model marries robot-centric decoding with a floor-wide contextual view. The option to decode along the robot axis or time axis lets us: (i)~encode an entire fleet snapshot in a single pass for fast, globally consistent action updates, and (ii)~decode long per-robot histories with causal masking and cached memories when extended temporal context is essential.

Together, RC and RF bracket the local-to-global spectrum in \DeepFleet, framing our study of spatiotemporal design trade-offs in large-scale multi-robot prediction.

%% file: sections/07_Image_Floor_Model.tex
\section{Image-Floor Model}
\label{sec:if_model}
The image-floor (IF) model is a floor-centric architecture that forecasts joint fleet evolution by treating the entire warehouse as a multi-channel image series.
\begin{figure}[t]
\centering
\includegraphics[width=0.99\linewidth]{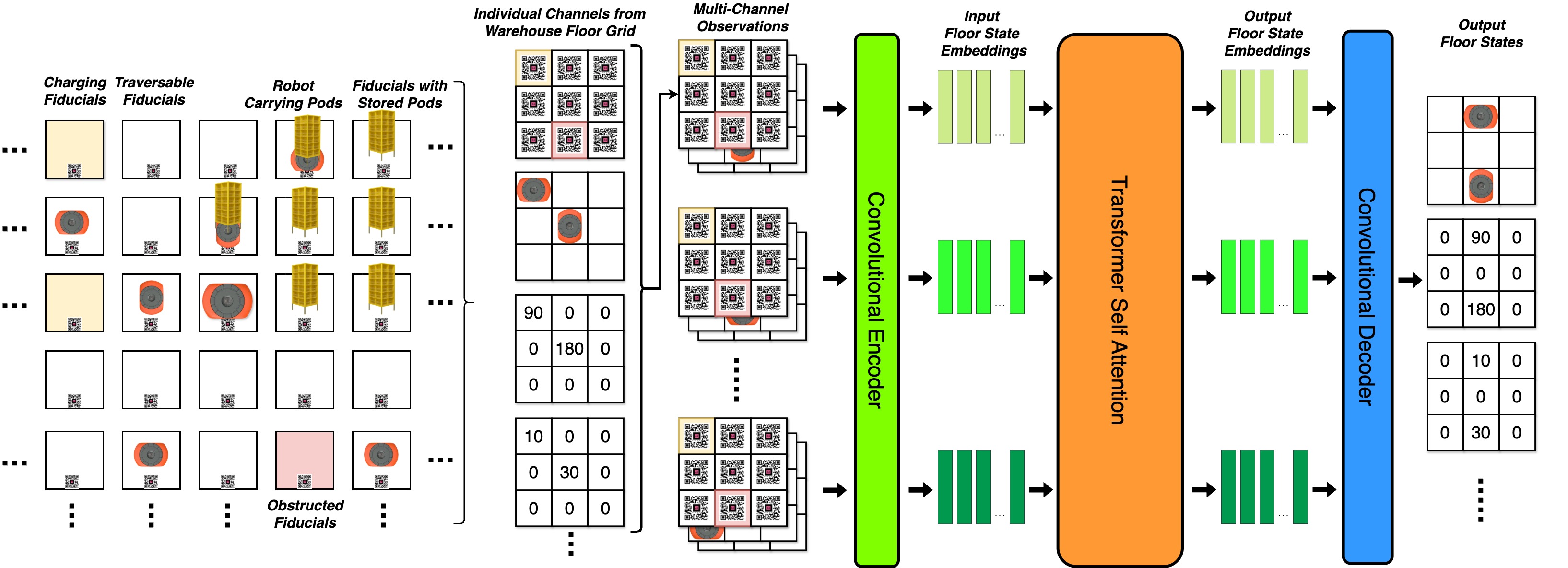}
\caption{
Image-Floor model architecture. The floor is rasterized into multi-channel image tensors with static and dynamic features. The spatial encoder uses convolutional blocks to downsample the input, the temporal transformer processes sequences of latent maps across time steps, and the spatial decoder upsamples back to full resolution with skip connections. Per-cell heads output motion and state predictions for each vertex.
}
\label{fig:if_architecture}
\end{figure}
Within the \DeepFleet design matrix it occupies the quadrant of fixed-time, synchronous predictors that enjoy full spatial context while modeling time autoregressively.  
Like video prediction models \cite{xing2024survey, zhou2024survey}, the IF model consumes a window of past floor frames and extrapolates the next frame; repeating this step yields a multi-step forecast. This dense, image-level strategy complements the sparse view of the RC model and the tokenized views of the RF and GF models by enabling direct pixel-wise reasoning over topology, traffic density, and local
congestion patterns that emerge only when the floor is viewed as a whole. Because every robot is updated simultaneously, the model can exploit GPU-friendly batched convolutions and achieve constant latency---independent of fleet size---during both training and inference.

At each time step the floor is rasterized into an $H\times W\times C$ tensor, where $(H,W)$ matches the floor size and $C$ denotes feature channels. Static channels encode immutable topology information, while dynamic channels encode per-robot quantities such as occupancy and robot state. Each grid cell can host up to two robots; when this occurs, their features are concatenated along the channel dimension and the excess left blank. To keep the channel count moderate we partition features into semantic blocks (topology, kinematics, mission cues) and normalize each block separately, yielding well-behaved statistics across sites of different sizes and traffic patterns. Finally, a learned positional embedding map $\mathbf P\in\mathbb R^{H\times W\times d_e}$ is added to every feature stack, injecting absolute floor coordinates so that the convolutional encoder can distinguish visually similar but physically distant cells.

\subsection{Model Framework}
The IF network follows an
encode$\rightarrow$temporally process$\rightarrow$decode pattern:
\begin{enumerate}[leftmargin=*,nosep,itemsep=2pt]
\item \textbf{Spatial Encoder.} A stack of convolutional blocks with residual skip paths downsamples the input image by powers of two, producing a latent map $\mathbf z_t\in\mathbb R^{\frac{H}{8}\times\frac{W}{8}\times d}$ that summarizes local context within an $8\times8$ window.  Group normalization and SiLU activations stabilize training, while a $1\times1$ projection aligns channel depth with the transformer's hidden size.

\item \textbf{Temporal Transformer.} For a history window $(\mathbf z_{t-K},\ldots,\mathbf z_t)$ we flatten each latent map into a sequence of $N_\mathrm{patch}$ tokens and feed the resulting $(K+1)\times N_\mathrm{patch}$ token stream into a decoder-only transformer with causal masking. Multi-head self-attention thus mixes information \emph{both} across time and across spatial patches, capturing complex interactions such as congested crossroads or queue spill-back. Rotary positional embeddings preserve temporal order without adding quadratic cost.

\item \textbf{Spatial Decoder.} The transformer's output tokens are reshaped back to a latent map $\tilde{\mathbf z}_{t+1}$, then upsampled through a mirrored convolutional decoder with skip connections from matching encoder stages. These lateral shortcuts re-inject fine-scale geometry that may be lost in the bottleneck, improving localization of narrow aisles and corner cells.

\item \textbf{Per-Cell Heads.} Two light MLP heads are applied per pixel: a regression head outputs $\Delta \mathbf{x},\Delta \mathbf{y}$ (Manhattan distance) for motion, and a classification head predicts laden flag \& orientation class. A gated fusion module allows the heads to share low-level features while learning task-specific cues.
\end{enumerate}

\subsection{Training Objective}
At its core, the IF model solves a multi-task learning problem: it must (i)~regress continuous motion offsets for every active robot cell, and (ii)~classify discrete state attributes (load status and heading) for those same cells. We therefore adopt a composite loss that couples a mean-squared-error term (for kinematics) with a focal cross-entropy term (for categorical state), allowing the network to share early features while giving each task a dedicated error signal. Formally, Let ${\cal S}$ be the set of grid cells that contain at least one robot in the \emph{ground-truth} target frame. For each such cell $p\in{\cal S}$ we denote the true motion vector by $\mathbf m_p=(\Delta x,\Delta y)$ and the model's prediction by $\hat{\mathbf m}_p$. Similarly, $\mathbf y^c_p$ and $\hat{\mathbf y}^c_p$ are the one-hot target and post-softmax probabilities for categorical class $c\!\in\!\{\text{load},\text{heading}\}$. The joint loss is
\begin{align}
\mathcal L_{\text{IF}}
  \;=\;
  \lambda_{\text{reg}}\;
  \frac{1}{|{\cal S}|}
  \sum_{p\in{\cal S}}
     \bigl\lVert\hat{\mathbf m}_p-\mathbf m_p\bigr\rVert_2^{\;2}
  \;+\;
  \lambda_{\text{cls}}
  \sum_{c\in\{\text{load},\text{heading}\}}
     \mathrm{FL}\!\bigl(\hat{\mathbf y}^c,\mathbf y^c\bigr),
  \label{eq:if_loss}
\end{align}
where $\mathrm{FL}(\cdot)$ is the focal loss $\mathrm{FL}(p_t)=\alpha_t(1-p_t)^{\gamma}\log p_t$ with $(\alpha_t,\gamma)=(0.25,2)$. The first term is a per-cell MSE that drives accurate regression of $\Delta x$ and $\Delta y$ displacements; the second term down-weights easy negatives so the classifier focuses on rarer events such as rotations or load transitions. Sampling only the active set ${\cal S}$ prevents static background pixels from overwhelming the motion loss, while still letting the decoder see full images during the forward pass.

By synthesizing spatially dense encodings with sequence-level attention, the IF model offers a GPU-efficient framework for whole-floor forecasting. It captures congestion waves, lane-level interactions, and topology-conditioned motion patterns that are difficult to express in token-based models. 

\subsection{Inference Roll-out}
Given the current floor tensor $\mathbf I_t$, the IF model predicts $\hat{\mathbf I}_{t+1}$ and feeds this prediction back as input, unrolling for $H$ future steps:

\begin{enumerate}[leftmargin=*,nosep,itemsep=2pt]
\item \textbf{Identity Propagation.} Each robot ID channel is translated to its predicted cell; ties (two robots competing for the same cell) are broken by highest motion confidence, mirroring live controller arbitration.
\item \textbf{Topology Re-injection.} Static topology channels are never predicted; instead they are copied verbatim into every autoregressive input so the model focuses capacity on dynamic state evolution.
\item \textbf{Latency.} Because convolution and per-cell heads run fully in parallel, inference time depends only on floor resolution $(H,W)$, not on robot count~$N$. On a modern GPU the model delivers $\sim$20\,ms step latency for a $256\times256$ grid---fast enough for 1Hz to 5Hz controller loops.
\end{enumerate}

%% file: sections/08_Graph_Floor_Model.tex
\section{Graph-Floor Model}
\label{sec:gf_model}

The graph-floor (GF) model targets the remaining cell of the design matrix, global spatial reach with synchronous temporal updates and action-based outputs just like the IF model. In addition, it uses action prediction for each robot on the floor similar to RC and RF models. Unlike the RF model, which processes a flattened token sequence, the GF model embeds the entire warehouse in a spatiotemporal graph ${G}_T = ({V}_T, {E}_T)$ and preserves this graph structure through every temporal layer, enabling it to naturally encode topological constraints while maintaining global spatial awareness and coordinated action prediction across the fleet.

\begin{figure}[t]
\centering
\includegraphics[width=0.98\linewidth]{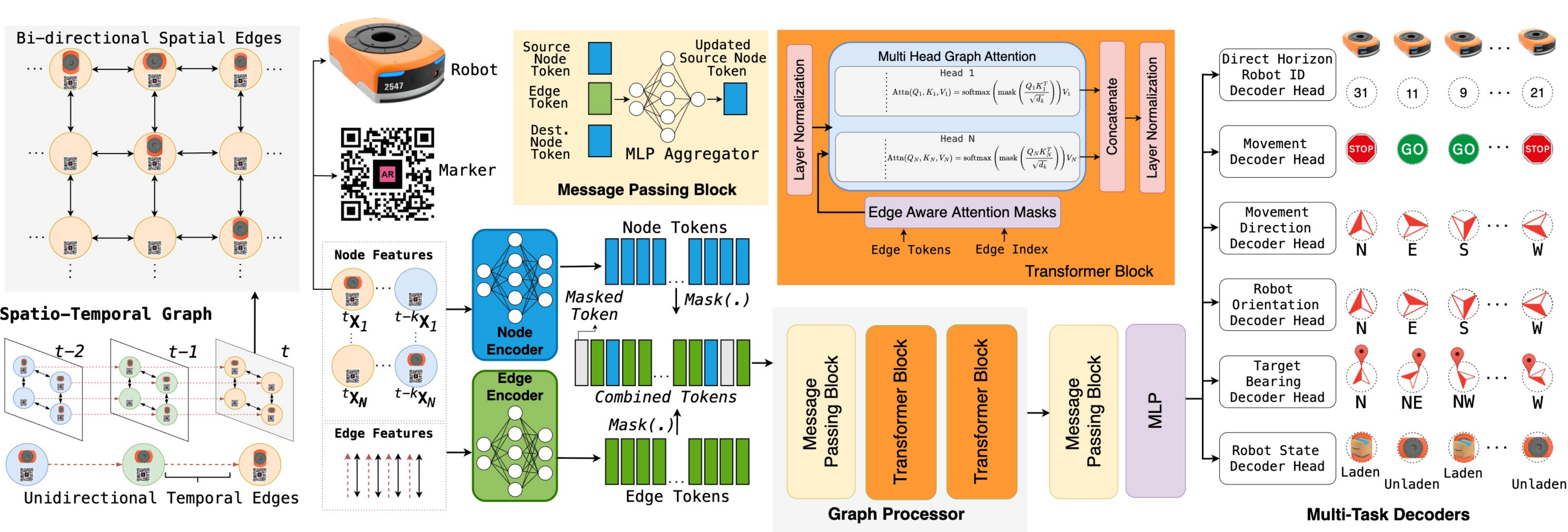}
\caption{
Graph-Floor (GF) model architecture. The model processes spatiotemporal data through (1)~a graph encoder that embeds node and edge features into a unified representation space; (2)~a graph processor consisting of stacked blocks with message-passing layers and edge-conditioned self-attention mechanisms; and (3)~a multi-task decoder that outputs robot movement action and state predictions.
}
\label{fig:GF_architecture}
\end{figure}
This inductive bias yields permutation-invariant reasoning and naturally encodes topological constraints (e.g., one-way aisles, choke-points, blocked edges). From the final node embeddings the network jointly outputs, for every robot, a discrete action and a continuous state---so that a single forward pass delivers the complete control vector required to advance the global floor state.

The GF model directly represents the physical environment using the graph $G = (V, E)$ introduced in Section~\ref{sec:system_overview}. This spatial representation is extended into the temporal domain by constructing a spatiotemporal graph ${G}_T = ({V}_T, {E}_T)$ that models the warehouse dynamics over time. The spatiotemporal node set ${V}_T = {V} \times \{1,...,T\}$ represents each vertex in $G$ at a discrete timestep, where a node $(v_i, t)$ captures both the static attributes of vertex $v_i$ and the dynamic attributes of any robot occupying that vertex at time $t$ (using a zero vector when unoccupied). The spatiotemporal edge set ${E}_T$ is partitioned into two components: the spatial edges, ${E}_\text{spatial} = \{((v_i, t), (v_j, t)) \mid (v_i, v_j) \in {E}, t \in \{1,...,T\}\}$ which preserve the spatial connectivity structure at each timestep; and the temporal edges, ${E}_\text{temporal} = \{((v_i, t), (v_i, t+1)) \mid v_i \in {V}, t \in \{1,...,T-1\}\}$ which connect each vertex to itself at adjacent timesteps, enabling the model to capture temporal dependencies. 
This augmented graph structure captures both temporal continuity and spatial proximity (between robots and warehouse topology), enabling comprehensive modeling of the warehouse's spatiotemporal dynamics.

\subsection{Model Framework}
The GF network $\mathcal F_{\theta}$ follows an
encode$\rightarrow$process$\rightarrow$decode pattern where raw node/edge features $\mathbf X$ are first embedded using an encoder, then enhanced with neighborhood context via processing block consisting of graph message-passing and transformer blocks to yield contextual node states $\mathbf H$, and finally multi task-heads decode each node's embeddings into task specific outputs.

\subsubsection{Graph encoder}
The graph encoder transforms raw node and edge features into a unified embedding space. For each node, the raw feature vector \(\mathbf{x}_i\) is processed via a node encoder $\mathcal{F}_{\theta_{E}^{n}}$ to produce a node token embedding as $\mathbf{t}_i = \mathcal{F}_{\theta_{E}^{n}}(\mathbf{x}_i) \in \mathbb{R}^{d_{\text{emb}}}$. Similarly, each edge's raw feature vector \(\mathbf{x}_{ij} \in \mathbb{R}^{d_{\text{e,in}}}\) is embedded using an edge encoder $\mathcal{F}_{\theta_{E}^{e}}$ into an edge token $\mathbf{t}_{ij} = \mathcal{F}_{\theta_{E}^{e}}(\mathbf{x}_{ij}) \in \mathbb{R}^{d_{\text{emb}}}$. These encoded representations preserve essential spatial, temporal, and relational nuances, and by embedding them in a high-dimensional latent space, they provide a robust foundation that enables subsequent processes to learn intricate fleet movement interactions.

\subsubsection{Graph processor}
The processor stacks $L$ identical graph blocks. Block $l$ receives token set $\mathbf T^{(l-1)}$ and returns $\mathbf T^{(l)}$ via message-passing layers and edge-conditioned self-attention.
After $L$ blocks we obtain the contextual node embeddings $\mathbf H=\{\mathbf h_{i,t}\}$ that carry both local and global spatiotemporal information.

\paragraph{Message-Passing Layers.} At each layer $l$ every edge $(v_i,v_j)$ emits a message
\[ \mathbf m^{(l)}_{ij}= \mathcal F^{\text{msg}(l)}_{\theta_P}  \bigl(\mathbf t^{(l-1)}_i,\,\mathbf t^{(l-1)}_j,\,\mathbf t_{ij}\bigr).
\] 
Node $v_j$ then aggregates messages from its neighbors $\mathcal N(j)$ and updates its embedding with another MLP:
\[ \mathbf t^{(l)}_j= \mathcal F^{\text{upd}(l)}_{\theta_P}\! \Bigl(\mathbf t^{(l-1)}_j \,\oplus\! \sum_{i\in\mathcal N(j)}\mathbf m^{(l)}_{ij}\Bigr). \]
Stacking these MLP-based message-passing blocks propagates local information throughout the graph while keeping computation linear in the number of edges.

\paragraph{Edge-Conditioned Self-Attention.} To capture long-range dependencies that pure message passing misses, every block also applies an edge-conditioned transformer to the current node states.  Concretely, for a query node $v_i$ we form key-value pairs $(\mathbf k_{ij},\mathbf v_{ij})$ by concatenating the neighbor embedding $\mathbf t^{(l-1)}_j$ with the learned edge token $\mathbf t_{ij}$: $\mathbf k_{ij}=W_K[\mathbf t^{(l-1)}_j\!\oplus\!\mathbf t_{ij}]$ and $\mathbf v_{ij}=W_V[\mathbf t^{(l-1)}_j\!\oplus\!\mathbf t_{ij}]$. The attention score $e_{ij}= \bigl\langle W_Q\mathbf t^{(l-1)}_i,\;\mathbf k_{ij}\bigr\rangle/\sqrt{d}$ therefore depends not only on the neighbor's state but also on the relationship between the two vertices (e.g., their distance). A softmax over $\mathcal N(i)$ produces weights $\alpha_{ij}$ and the updated node embedding is the weighted sum $\mathbf t^{(l)}_i=\sum_{j\in\mathcal N(i)}\alpha_{ij}\mathbf v_{ij}$. Because edge features are injected directly into the key-value pipeline, this module preserves permutation invariance while allowing the network to reason about asymmetric or time-varying constraints, giving a single block the receptive field of the whole floor without quadratic cost over all $M$ vertices.

\subsubsection{Multi-task decoder} In our multi-task decoder, we decompose the overall prediction problem into several interconnected sub-tasks, each handled by a dedicated prediction head that shares underlying representations. Specifically, refined node representations \(\mathbf{h}_i \in \mathbb{R}^d\) are mapped to task-specific outputs via $\mathbf{y}_k = \mathcal{F}_{\theta^k_{D}}(\mathbf{h}_i), \quad k \in \{1, \ldots, K\}$, where the index \(k\) denotes each of the \(K\) specialized prediction heads corresponding to distinct prediction tasks. The tasks are as follows:
\begin{itemize}[leftmargin=*,nosep,itemsep=2pt]
\item \textbf{Direct Horizon Robot ID Prediction:} predicts robot identities across vertices over a horizon of \(H\) timesteps for consistent tracking in a single model pass. 

\item \textbf{Movement Prediction:} determines if a robot should move or wait based on surrounding context.

\item \textbf{Movement Direction Prediction:} forecasts the robot's movement direction.

\item \textbf{Orientation Prediction:} predicts the robot's heading for efficient navigation.

\item \textbf{Robot State Prediction:} classifies the operational state (laden or unladen) to based on the input history of past states.

\item \textbf{Target Bearing Prediction:} outputs a continuous 2D vector representing the normalized direction toward the target.
\end{itemize}
The interconnection among these tasks allows the model to leverage complementary information, thereby enhancing the overall prediction of fleet movement dynamics.

\subsection{Training Objective}
We supervise the network with a single loss that is the weighted sum of task-specific terms:
\begin{equation}\label{objective_gf}
\mathcal L_{\text{total}}
    = w_{\text{ID}}\,\mathcal L_{\text{ID}}
    + w_{\text{move}}\!\left(\mathcal L_{\text{move}}
                       +  \mathcal L_{\text{dir}}\right)
    + w_{\text{attr}}\,\mathcal L_{\text{attr}}
\end{equation}
where  
$\mathcal L_{\text{ID}}$,
$\mathcal L_{\text{move}}$, and
$\mathcal L_{\text{dir}}$, are cross-entropy terms for
robot identity tracking, move/wait, and direction, respectively, while  
$\mathcal L_{\text{attr}}$ is a cosine loss over continuous attributes (heading, load state, target bearing). The scalars $w_{\ast}$ balance the relative importance of these components and are tuned on a validation set. Gradients flow through the entire encode--process--decode pipeline, allowing the model to discover the spatial and temporal cues that best explain future fleet behavior.

In summary, the GF model further explores the \DeepFleet spectrum by pairing global spatial awareness with long, event-driven temporal reasoning. A graph encoder embeds nodes and edges; message-passing plus edge-aware self-attention propagate context over arbitrary distances; and a multi-task decoder jointly outputs move/rotate actions, heading angles, and laden state for all the robots.   
Together with the local RC and snapshot-based RF baselines, the GF model frames a comprehensive set of inductive biases for studying large-scale multi-robot prediction.

\subsection{Inference Roll-out}
During inference, the GF model predicts future states in an autoregressive loop. At each time step $\tau$ we perform three steps:
\begin{enumerate}[leftmargin=1.4em,itemsep=2pt]
\item \textbf{Forward pass.} Feed the current spatiotemporal graph slice $\mathbf S_\tau$ through the network $\mathcal F_\theta$ to obtain, for every robot node, the  robot movement action $\hat a_\tau$, the target heading $\hat\theta_\tau$, and the load status $\hat\ell_\tau$.

\item \textbf{State transition.} Apply the deterministic floor dynamics $f_{\text{floor}}\!:\!(\hat a_\tau,\hat\theta_\tau,\hat\ell_\tau) \mapsto\mathbf S_{\tau+1}$: robots that move update their vertex, pose, and load status; robots that wait simply copy their previous state.

\item \textbf{Collision arbitration.} If two robots claim the same destination vertex, the request that has higher confidence is honored and the second robot is rolled back to its prior pose (mirroring the first-come-first-serve rule used in the RC model).
\end{enumerate}
Because the processor caches key-value memories from the
preceding step, each prediction step costs
$O(|V|+|E|)$, which is linear in the warehouse size. This loop is repeated until the desired forecast horizon is reached, yielding a consistent, step-by-step simulation of future fleet behavior.

%% file: sections/09_conclusion.tex
\section{Conclusions}
\label{sec:conclusion}
We presented \DeepFleet, a collection of four distinct architectures for neural networks that are pre-trained to predict large-scale mobile robot fleet movement.
The four models cover the design space for multi-robot foundation models, spanning across different types of priors, representations of static and dynamic features, and neural families.
We trained the models on millions of robot hours of production data from Amazon warehouses, and evaluated them on their ability to predict future states autoregressively. 

Our empirical analysis revealed key learnings for the design of multi-agent foundation models. First, we learned that next action prediction outperforms floor state prediction: both the RC model, which predicts the next actions of robots based on their state and that of their immediate environment, and the RF model, which also explicitly cross-attends floor and robot features, outperformed our floor-based models. We also learned that using convolutional encoding over an image-like representation of the floor and robots performs poorly, even at comparable model sizes. We suspect this representation does not provide adequate inductive bias to model robot interactions, which occur at a pixel level. Our results also show that explicitly modeling spatiotemporal relations between robots and floor (the GF model) is able to predict robot interactions, even at significantly lower parameter counts.

We performed a scaling study on two promising models---RC, the most accurate predictor of next state, and GF, the smallest model with comparable performance---demonstrating that scaling model size, compute budget, and dataset size improve performance. While the RC model needs further experiments to fully characterize its scaling, we were able to derive a scaling law for the GF model that can be extrapolated to predict the optimal mix of model and dataset size. These results will be used to further improve the models as we work to develop downstream applications in our warehouses.

%% file: sections/A_Contributors.tex
\section{List of Contributors}
\label{sec:authors}
Please address correspondence to \texttt{deepfleet@amazon.com}.

{\bf Scaling experiments and writing lead}\\
{\em Ameya Agaskar}\\
{\em Sriram Siva}

{\bf DeepFleet model design and development}\\
{\em William Pickering (Robot-Centric model)}\\
{\em Kyle O'Brien (Robot-Floor Cross Attention model)}\\
{\em Charles Kekeh (Image-Based Floor-Centric model)}\\
{\em Sriram Siva (Graph-Based Floor-Centric model)}\\
{\em Alexandre Ormiga Galvao Barbosa (Multiple)}

{\bf Comparative empirical evaluation}\\
{\em Ang Li}\\
{\em Brianna Gallo Sarker}\\
{\em Alicia Chua}

{\bf Data pipelines and infrastructure}\\
{\em Mayur Nemade}\\
{\em Charun Thattai}\\
{\em Jiaming Di}\\
{\em Isaac Iyengar}\\
{\em Ramya Dharoor}\\
{\em Dino Kirouani}

{\bf Project management}\\
{\em Jimmy Erskine}\\
{\em Tamir Hegazy}

{\bf Scientific guidance}\\
{\em Scott Niekum (Amazon Scholar)}\\
{\em Usman A. Khan (Amazon Scholar)}

{\bf PI and research program lead}\\
{\em Federico Pecora}\\
{\em Joseph W.~Durham}

\section*{Acknowledgments}
The authors wish to thank Zhe Chen for helpful discussions on multi-agent pathfinding (MAPF).